\documentclass[conference]{IEEEtran}
\usepackage{graphicx,times,amsfonts,amssymb,amsmath,multirow,subfigure} 
\usepackage{algorithmic,algorithm,url}

\usepackage{tikz}
\usepackage{textcomp}
\usepackage{hyperref}
\usepackage{lipsum}

\newcommand\copyrighttext{%
  \footnotesize \textcopyright 2017 IEEE. Personal use of this material is permitted.
  Permission from IEEE must be obtained for all other uses, in any current or future
  media, including reprinting/republishing this material for advertising or promotional
  purposes, creating new collective works, for resale or redistribution to servers or
  lists, or reuse of any copyrighted component of this work in other works.\\
  \emph{To appear in the Proceedings of the 2017  IEEE Symposium Series on Computational Intelligence (IEEE SSCI 2017)}}
\newcommand\copyrightnotice{%
\begin{tikzpicture}[remember picture,overlay]
\node[anchor=north,yshift=-10pt] at (current page.north) {\fbox{\parbox{\dimexpr\textwidth-\fboxsep-\fboxrule\relax}{\copyrighttext}}};
\end{tikzpicture}%
}

\IEEEoverridecommandlockouts    

\begin{document}


\title{Hidden Tree Markov Networks: Deep and Wide Learning for Structured Data}
\author{\IEEEauthorblockN{Davide Bacciu}
\IEEEauthorblockA{Dipartimento di Informatica, Universit\`a di Pisa\\
Largo B. Pontecorvo 3, Pisa, Italy\\
Email: bacciu@di.unipi.it}
}


\maketitle
\copyrightnotice

\begin{abstract}
The paper introduces the Hidden Tree Markov Network (HTN), a neuro-probabilistic hybrid fusing the representation power of generative models for trees with the incremental and discriminative learning capabilities of neural networks. We put forward a modular architecture in which multiple generative models of limited complexity are trained to learn structural feature detectors whose outputs are then combined and integrated by neural layers at a later stage. In this respect, the model is both deep, thanks to the unfolding of the generative models on the input structures, as well as wide, given the potentially large number of generative modules that can be trained in parallel. Experimental results show that the proposed approach can outperform state-of-the-art syntactic kernels as well as generative kernels built on the same probabilistic model as the HTN.
\end{abstract}

\IEEEpeerreviewmaketitle

\section{Introduction} \label{sect:intro}
The deep learning revolution is strongly rooted on the ability of efficiently learning informative neural representations that effectively inform the predictor part of the model, starting from large scale, complex, and often noisy, data. This has produced breakthrough performances in several fields of computer science, such as machine vision, speech recognition and natural language understanding. These specific application fields are often associated with data of very specialized nature, such as images for machine vision and sequences for speech and text processing.

Recently, it can be noted an increasing attention of the deep learning community on a complex type of data that naturally describes hierarchical information, that is tree-structured data. Trees can be thought of as compound objects made by atomic entities, represented by the knowledge encoded in the node labels, that are bound together by hierarchical relationships, represented by the tree edges. With this interpretation in mind, one can clearly note how an effective representation of the information associated to a node cannot be determined by considering the node in isolation. Rather, it needs to take into consideration the surrounding context, represented by the nodes a target vertex it is linked to. A key challenge is that, within such a scenario, data samples are trees of varying structure and size. Hence the learning model needs to be adaptive with respect to such variations.

Deep learning approaches to tree-structured data have developed mostly as variations of the Long Short Term Memory (LSTM)\cite{TaiSM15}, where the original cell architecture designed to process sequences has been extended to model different parsing directions (recall that a tree can be visited top-down or bottom-up producing different node neighborhoods) as well as larger contextual dependencies (e.g. from the set of children of a node instead of the single predecessor of the sequence). In fact, most of the deep LSTM-based approaches are basically implementing a specific instance of a very general approach to tree structured data processing proposed in the late nineties by \cite{frasconi1998general}.

Inspired by the work on relative density networks \cite{brown2002relative}, in this paper, we take a completely different approach by proposing a neuro-probabilistic hybrid model which allows to learn effective encodings of discriminative structural knowledge using generative tree models immersed in a neural architecture, that is then used to perform the final predictive task (i.e. tree classification for the sake of this work). The model is strongly rooted on characterizing concepts of \emph{deep} learning, such as modularity of the network and a profoundly layered architecture that organizes information into a hierarchical form. This is further extended in a \emph{wide} sense, as it allows to concatenate in parallel a large number of such generative structural feature detectors. These are intended do be of small complexity (in terms of number of parameters and consequently computational complexity), allowing to have many small, easily trainable, structural detectors in place of a huge monolithic generative model which is generally quite harder to train and to perform inference onto. We show what architectural choices need to be taken in order to make such feature detectors tuned to different structural properties and we experimentally assess the effect of discriminatively training the generative models within a neural architecture. The results highlight how such an approach can outperform other methods for imbuing discriminative power into generative models, e.g. through the definition of generative kernels \cite{genKernel}. The modular structure of the network allows to apply a wide range of performance and training optimization tricks which might prove effective when dealing with large scale problem. For instance, the single generative models can be pre-trained in an unsupervised fashion before being embedded in the neural architecture, where gradient descent can then be used to refine the parameters based on supervised information. On the other hand, the modularity of the network allows to straightforwardly parallelize the computations in the generative models, for instance by resorting to minibatching techniques for computational efficiency.

The remainder of the paper is organized as follows: Section \ref{sect:background} reviews the background on adaptive processing of tree-structured data, placing particular focus on the formalization of the generative models underlying the proposed approach. Section \ref{sect:approach} introduces the proposed model which we name \emph{Hidden Tree Markov Network} (HTN). Section \ref{sect:expcomp} provides an experimental assessment of the HTN, comparing it with state-of-the-art kernels for trees, while Section \ref{sect:conclude} concludes the paper.

\section{Background} \label{sect:background}

\subsection{Learning with Tree Structured Data} \label{sect:notation}
The paper deals with models that can learn predictive tasks from labelled datasets whose constituents are tree-structured samples whose size and connectivity vary among data points. For the purpose of this paper, we will consider classifications tasks where we want to associate an input tree to a class label. To formalize the notation used throughout the paper, we consider a dataset  $\mathcal{D} = \{\mathbf{x}^1,\dots,\mathbf{x}^N\}$ of $N$ labeled rooted tree where $\mathbf{x}^{n}$ is a connected acyclic graph consisting of a set of nodes $\mathcal{U}_{n} = \{1,\dots,U_n\}$ such that a single vertex is denoted as the \emph{root} and any two nodes are connected by exactly one simple path. The index $n$ is used to denote the $n$-th tree in a dataset of $N$ structures and will be omitted for notational simplicity when the context is clear. The term $u \in \mathcal{U}_{n}$ is used to denote a generic node of $\mathbf{x}^{n}$, whose direct ancestor is called  \emph{parent} and it is denoted as $pa(u)$. By definition, each node $u$ has at maximum one parent, but it can have a variable number of direct descendants (\emph{children}), such that the $l$-th child of node $u$ is denoted as $ch_l(u)$. Note that here we assume trees to have a finite maximum \emph{outdegree} $L$, i.e. the maximum number of children of a node. A node without children is called \emph{leaf} and the set of leaves of the $n$-th tree is denoted as $\mathcal{LF}_{n}$. Finally, each vertex $u$ in the tree is associated with a label $x_u$ which is a $d$-dimensional vector.

The literature on adaptive processing of tree-structures is quite rich, comprising both kernel-based, generative and neural approaches. A key foundational work providing a reference framework for the processing of tree structured data is \cite{frasconi1998general}, where it is proposed a unifying approach to deal with such data both from a neural and a probabilistic perspective. More recently, the deep learning wave has re-discovered the use of recursive neural networks for the processing of trees. In particular, \cite{Socher2013} proposed a recursive neural network which basically implemented a specialized version of the general framework in \cite{frasconi1998general} within a parse tree classification task for emotion recognition. From this work, several other followed typically proposing recursive variants of the Long Short Term Memory (LSTM), e.g. see the bottom-up LSTM in \cite{TaiSM15} and the top-down LSTM in \cite{Lapata2016}.

An alternative approach to learning with structured data is put forward by using kernel functions to define similarity measures on trees upon which support vector machines are built to solve classification/regression problems. Several kernel functions have been proposed in the past-years to deal with structured data: an early survey is available in \cite{DBLP:journals/sigkdd/Gartner03}. Many of these are of syntactic type, that is a class of tree kernels where the degree of matching between two trees is determined by counting the number of common substructures among the trees \cite{sst}. The various approaches in literature basically differentiate by the way they identify the composing substructures and by how they weigh the structural matches. The Subset Tree kernel (SST) by \cite{sst}, for instance, counts the number of matching proper subtrees, while the Subtree kernel (ST) \cite{Vishwanathan2003} restricts to matching only complete subtrees for computational efficiency.  Elastic tree kernels \cite{elastic2002}, on the other hand, allow matching nodes with different labels and matching between substructures built by combining subtrees with their descendants.  The Partial Tree kernel (PT) \cite{Moschitti2006} relaxes SST to allow partial productions of the parse-tree grammar, basically allowing to perform partial matching between subtrees at the cost of an increased computational complexity. A recent comparative analysis of syntactic tree kernels can be found in \cite{Shin2014}.

Another class of adaptive approaches for trees is based on the use of generative models based on a hidden Markov state formulation, that are briefly reviewed in the next section as they are the basic building block upon which we build our solution. In particular, this paper proposes an hybrid with respect to the approaches in literature. We fuse in the same architecture the ability of generative models in extracting descriptive probability distributions on tree spaces, with the flexibility, efficiency and incremental learning ability of neural approaches. Further, in the experimental analysis, we show how such an hybrid reaches higher classification performances than state-of-the-art kernels, both syntactical and built on the top of the same generative models used by our approach.

\subsection{Hidden Tree Markov Models} \label{sect:htmm}
Hidden Tree Markov Models (HTMM) allow modeling probability distributions over spaces of trees by generalizing the Hidden Markov Model (HMM) approach for the sequential domain, through learning of an hidden generative process for labeled trees regulated by hidden state variables modeling the structural context of a node and determining the emission of its label. In literature, we refer mainly to two types of generative processes associated to top-down (TD) \cite{hmt98,htmm02} and bottom-up (BU) \cite{DBLP:journals/tnn/BacciuMS12,DBLP:journals/tnn/BacciuMS13} parsing directions. The TD and BU approaches are characterized by different context propagation strategies that, in practice, induce different conditional independence relationships, leading to different representational capabilities. In the following we focus on the BU version of the HTMM, as it has been shown that the conditional dependence relationships introduced by the BU context allow, in general, to capture more discriminant details on the tree structures with respect to a TD context \cite{DBLP:journals/tnn/BacciuMS12}.

\begin{figure}[tb]
\includegraphics[width=.98\columnwidth]{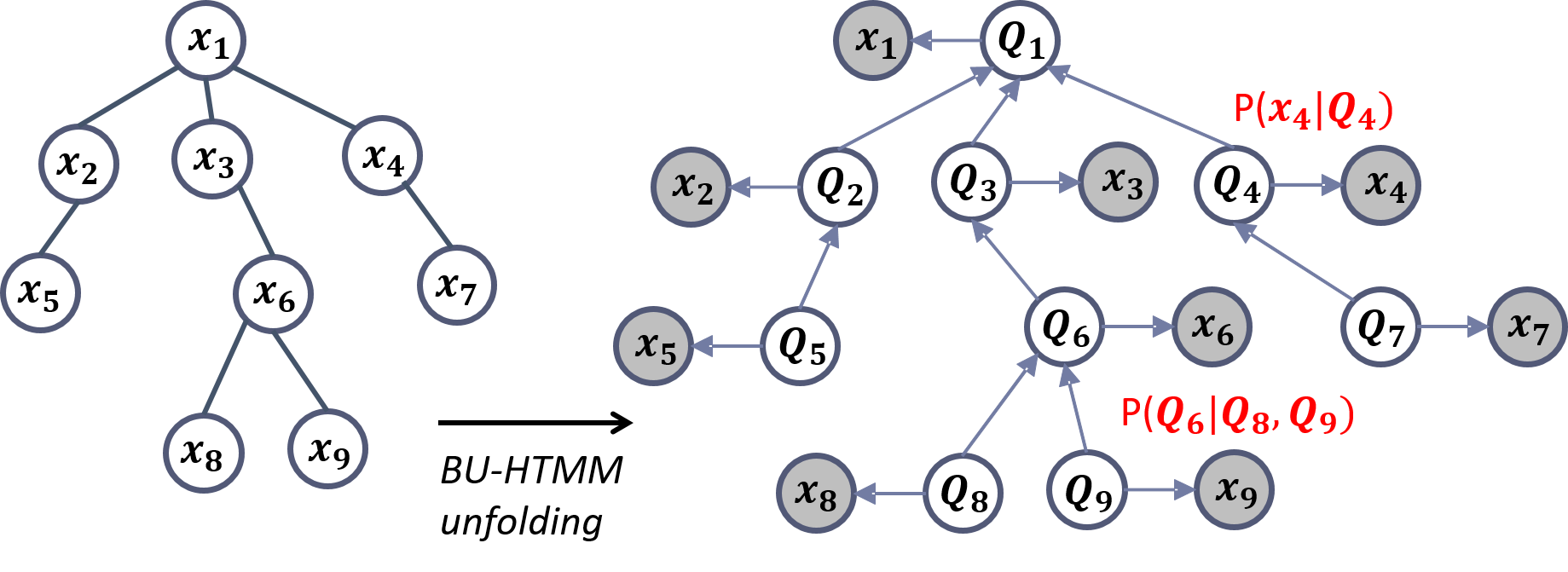}
\caption{Unfolding of a BU-HTMM on the structure of an example tree (on the left), generating the directed graphical model (on the right) comprising visible label nodes (shaded) as well as hidden Markov state nodes (empty). The figure shows an example of the emission distribution $P(x_4|Q_4)$ for node $4$ and of a transition distribution $P(Q_6|Q_8,Q_9)$ for node $6$. \label{fig:unfold}}
\end{figure}

The BU-HTMM \cite{DBLP:journals/tnn/BacciuMS12} defines a generative process propagating from the leaves to the root of the tree, which allows nodes to collect dependency information from each child subtree. The BU-HTMM implements a generative process that composes the child subtrees of each node in the tree in a recursive fashion. In practice, BU-HTMM is a recursive model whose associated graphical model is different for each tree and it is obtained by unfolding on the structure of the target tree as described in \cite{frasconi1998general}. Figure \ref{fig:unfold} shows an example of such an unfolding. To realize this, we rely on a set of hidden state variables associated with a state transition dynamics that follows the direction of the generative process. Specifically, an observed tree $\mathbf{x}^n$ is modeled by a set of hidden state variables $\{Q_1, \dots, Q_u, \dots\}$ following the same indexing as the observed nodes $u \in \mathcal{U}_n$ and assuming values on the discrete set of hidden states $\{1,\dots,C\}$. The direction of the generative process is then modeled by the \emph{state transition probability}
\begin{equation} \label{eq:transBU}
P(Q_u = i | Q_{ch_1(u)} = j_1, \dots, Q_{ch_L(u)} = j_L)
\end{equation}
assuming that each node $u$ is conditionally independent of the rest of the tree when the joint hidden state of its direct descendants $Q_{ch_l(u)} = j_l$ is observed.  To complete the specification of the model in Figure \ref{fig:unfold}, we assume that the label $x_u$ (continuous or discrete) of a node $u$ is completely specified by its hidden state $Q_u$ through the \emph{emission distribution} $P(x_u | Q_u = j)$.

The problem with the formulation in eq.~(\ref{eq:transBU}) is that it becomes computationally impractical for trees other than binary, since the size of the joint conditional transition distribution is order of $C^{L + 1}$, where $L$ is the node outdegree. In  \cite{DBLP:journals/tnn/BacciuMS12}, this has been addressed by introducing a scalable \emph{switching parent} approximation that factorizes (\ref{eq:transBU}) as a mixture of $L$ pairwise child-parent transitions. The resulting BU-HTMM joint distribution is
\begin{equation}\label{eq:BHTMM}
\begin{split}
    P &(\mathbf{x}^{n}, Q_1, \dots, Q_{U_{n}}) = \prod_{u' \in \mathcal{LF}_n} P(Q_{u'}) P(x_{u'} | Q_{u'}) \\
    &\prod_{u \in \mathcal{U}_n \setminus \mathcal{LF}_n} P(x_u | Q_u) \sum_{l = 1}^L P(S_u = l) P(Q_u | Q_{ch_{l}(u)})
\end{split}
\end{equation}
where we recall that $\mathcal{LF}_n$ denotes the set of leaves in tree $\mathbf{x}^{n}$ and where $P(Q_{u'})$ is their \emph{prior} state distribution (given that leaf nodes have no children). The summation term in eq.~(\ref{eq:BHTMM}) corresponds to the factorization of (\ref{eq:transBU}) using the switching parent \mbox{$S_u \in \{1,\dots,L\}$}. This is a latent variable, independent from $Q_{ch_{l}(u)}$, and whose distribution \mbox{$P(S_u = l)$} measures the influence of the $l$-th children on a state transition to node $u$.

Learning the parameters of the  BU-HTMM  is addressed as an Expectation-Maximization (EM) \cite{dempster1977maximum} problem applied to the logarithm likelihood obtained from (\ref{eq:BHTMM}) by marginalizing the unknown hidden state assignment and summing on all trees in the dataset. The resulting likelihood is as follows
\begin{equation}\label{eq:SPBHTMMLik}
\begin{split}
    \mathcal{L} & = \prod_{n=1}^N \sum_{i_1, \dots, i_{U_n}} \prod_{u' \in \mathcal{LF}_n} P(Q_{u'} = i_{u'}) P(x_{u'} | Q_{u'} = i_{u'}) \\
    & \ \  \times \prod_{u \in  \mathcal{U}_n \setminus \mathcal{LF}_n} P(x_u | Q_u = i_u) \\
    & \ \ \ \times \left\{ \sum_{l = 1}^{L} P(S_u = l) P(Q_u = i_u | Q_{ch_{l}(u)} = i_{ch_l(u)})\right\}.
\end{split}
\end{equation}
Details of the learning algorithm can be found in \cite{DBLP:journals/tnn/BacciuMS12}. In summary, this is a batch learning algorithm based on a two-stage iterative procedure which allows to efficiently compute a solution to the log-likelihood maximization problem. At the E-step, it estimates the posterior of the indicator variables introduced in the completed log-likelihood, while, at the M-Step, it exploits such posteriors to update the model parameters $\theta$.  Posterior estimation is the most critical part of the algorithm and can be efficiently computed by message passing upwards and downwards on the structure of the nodes' dependency graph \cite{DBLP:journals/tnn/BacciuMS12}. The exact parameterization of the generative model depends on a number of factors, including the stationariety assumption taken and the form of the state and emission distribution. However, these are often of multinomial type, hence model parameters are quite straightforwardly the members of such multinomial tables, resulting in constrained maximization of (\ref{eq:SPBHTMMLik}).

A typical problem which one wants to address with an HTMM is that of tree classification. In a generative setting, such a task is typically addressed by training a different probabilistic model for each class (i.e. using solely sample trees from the target class). Then, for a test tree, all class-specific models are queried and the tree is assigned to the class whose corresponding model has the highest likelihood to have generated it. Unfortunately, such a generative approach typically yields to poor classification performances, as the learning model is not supplied with information that might help it in discriminating the different classes \cite{DBLP:journals/tnn/BacciuMS12}. Other approaches exists for the problem, such as introducing class information into the probabilistic model and using EM to perform class inference as in \cite{DBLP:journals/ijon/BacciuMS13}. However, even such models cannot reach the classification performance of discriminative approaches and their use cannot be easily generalized to regression problems. On the other hand, the structural knowledge inferred by the HTMM models can be used to inform discriminative methods, e.g. by building kernels that exploit an underlying generative model. This is, for instance, the case of the Fisher tree kernel \cite{nicotra2007generative} or of a family of adaptive tree kernels built using similarity metrics on multisets of HTMM state information \cite{genKernel,icann12,ijcnn2011}.

In the following section, we show a different way to approach the problem of making generative tree models discriminative by plugging them into a hierarchical neural network, ultimately realizing a neuro-probabilistic hybrid model for tree-structured data. Differently from the kernel-based solution above, our approach does not separate the phase of generative model training (i.e. the EM for the HTMM) from that of learning the discriminative predictor (i.e. training the support vector model using the tree kernel).  Instead, the generative models are trained discriminatively altogether with the neurons performing the  network prediction.

\section{Hidden Tree Markov Network (HTN)} \label{sect:approach}

\subsection{The Model}
The predictive accuracy of generative models on classification tasks is limited by the generative training style of the EM algorithm which basically amounts to learning class-conditional distributions $P(\mathbf{x}|k)$, one for each class $k$ and from only positive examples of that class. The lack of information from negative class examples does not allow such models to develop a sufficiently discriminative representation of the class boundaries, resulting in poor classification performances. Several approaches have attempted addressing this aspect within the context of structured data processing. Input-driven generative models \cite{DBLP:journals/ijon/BacciuMS13}, for instance, provide a way of introducing discriminative information by training a single probabilistic model $P(k|\mathbf{x})$ that can predict an output class $k$ given the input structure $\mathbf{x}$ which parameterizes the model distributions. In the context of sequential data, the Alphanet \cite{alphanet} approach proposed instead to attach a layer of softmax neurons receiving input from the likelihood of class conditional HMMs. It showed how to train such an hybrid model by cross-entropy maximization, reaching superior classification performances with respect to purely generative HMM training. Later, \cite{brown2002relative} introduced relative density net, combining HMMs through additional layers of hidden comparator neurons trainable by backpropagation.

\begin{figure}[tb]
\includegraphics[width=.98\columnwidth]{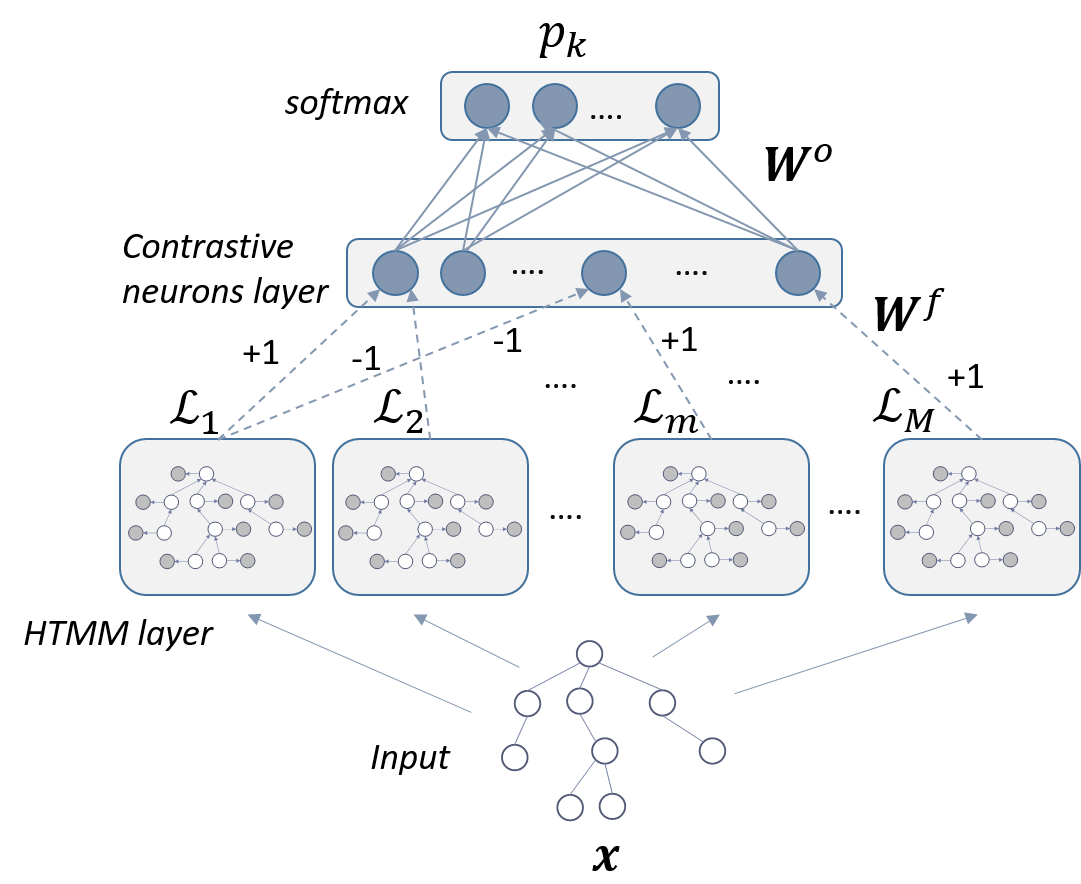}
\caption{Hidden Tree Network (HTN) comprising $M$ BU-HTMM whose likelihood $\mathcal{L}_m$ are compared through ${M \choose 2}$ contrastive neurons with fixed weights matrix $\mathbf{W}^{f}$ (dashed lines). Output neurons are softmax  (for tree classification) with adaptive weights $\mathbf{W}^{o}$ (solid lines). \label{fig:htn}}
\end{figure}

Motivated by the approaches above, we introduce the Hidden Tree Markov Network (HTN) for tree structured data, whose architecture is sketched in Fig.~\ref{fig:htn}. At the core of HTN sits a generative model for trees such as the BU-HTMM described in Section \ref{sect:htmm}, available in $M$ instances (all initially randomly initialized) represented as round boxes at the bottom of Fig.~\ref{fig:htn}. These BU-HTMMs unfold on the structure of the current input tree $\mathbf{x}$ and produce as a result a log-likelihood value $\mathcal{L}_m(\mathbf{x})$ for each module $m$, representing the probability of the current sample being generated by the $m$-th BU-HTMM. Such $\mathcal{L}_m(\mathbf{x})$ values are interpreted as outputs of the $M$ modules and combined pairwise by the successive layer of hidden contrastive neurons. For each pair of BU-HTMM $(m,r)$ with $m,r \in [1,M]$ and $m \neq r$, we have an hidden contrastive neuron $(m,r)$ whose activation is
\begin{equation}\label{eq:hidden}
  h(\mathbf{x})_{(m,r)} = \sigma(\mathcal{L}_m(\mathbf{x}) - \mathcal{L}_r(\mathbf{x}))
\end{equation}
where the activation function $\sigma(\cdot)$ is an hyperbolic tangent. Such a contrastive neuron acts as a comparator between two different generative models, whose likelihood is weighted by a fixed synaptic weight equal to either $+1$ or $-1$, depending on the module. By having models confronted pairwise with opposite signs, we expect them to develop different responses to the structures at training time. In other words, each BU-HTMM can be seen as a detector of some structural properties of the trees. By way of the comparator neurons, we expect different BU-HTMM models to become tuned to different structural features in the data. The contrastive neurons, in turn, can be interpreted as detectors of more abstract structural feature, following the deep learning idea of structuring knowledge in a hierarchical fashion from the more low level information at the input to more abstract features close to the network output.

Figure \ref{fig:htn} highlights how the contrastive neuron layer of HTN comprises ${M \choose 2}$ units that are sparsely connected to the $M$ generative models through a fixed weight matrix $\mathbf{W}^f$ whose element are in $\{-1,0,1\}$. The output of this layer is then combined through a fully connected layer of $K$ softmax neurons (one for each class $k$) with adaptive weight matrix $\mathbf{W}^o$. The softmax output $p_k(\mathbf{x}) = P(k | \mathbf{x})$ is computed as expected following
\begin{equation}\label{eq:soft}
    p_k(\mathbf{x}) = \frac{\exp{\left(in_{k}(\mathbf{x})\right)}}{\sum_{k'=1}^{C} \exp{\left(in_{k'}(\mathbf{x})\right)}}
\end{equation}
where
\begin{equation}\label{eq:softNet}
   in_{k}(\mathbf{x}) = \sum_{(m,r) \in \{{M \choose 2}\}} W_{(m,r)k}^o h(\mathbf{x})_{(m,r)}.
\end{equation}
The term $(m,r)$ in the argument of the sum in eq.~(\ref{eq:softNet}) denotes the hidden contrastive neuron connected to BU-HTMM modules $m$ and $r$. Hence the sum in eq.~(\ref{eq:softNet}) runs on all the ${M \choose 2}$ hidden neurons.

The HTN model is a neuro-probabilistic hybrid that is \emph{deep}, given that it comprises a level of recursive models whose depth grows with the input structure (as it happens with recursive neural networks). At the same time, we expect the HTN to be \emph{wide}, since it allows to have a large number $M$ of structural feature detectors of small complexity (i.e. number of hidden states $C$), in place of having fewer but very large HTMMs which are generally harder to train.



\subsection{Model Parameterization and Training}
Training of the HTN model for tree classification can be addressed by gradient descent minimization of  the cross-entropy loss, given an input tree $\mathbf{x}$, i.e.
\begin{equation}\label{eq:loss}
  L(\mathbf{x}|\theta) = - \sum_{k=1}^{K} d_k \log p_k(\mathbf{x}|\theta)
\end{equation}
where $d_k$ is the ground truth binary variable such that $p_k=1$ when $\mathbf{x}$ is of class $k$ and it is zero otherwise. In eq.~(\ref{eq:loss}) it has been introduced the term $\theta$ to explicitly denote the dependencies on the model parameters $\theta$. Cross-entropy minimization can be obtained by taking derivatives of (\ref{eq:loss}) with respect to the parameters $\theta$, that include both the adaptive output weights $\mathbf{W}^o$ as well as the parameters of the distributions of the BU-HTMM models. While the first is an unconstrained optimization problem, the second deals with an optimization with sum-to-one constraints due to the nature of the BU-HTMM parameters, which are essentially elements of multinomial distributions (i.e. prior, transition and emission probabilities). Although there exist approaches to solve such a constrained optimization problem in the field of hidden Markov models (see \cite{hmmTrain} for a survey), it is generally more effective to introduce a reparametrization of the Markov model such that learning can be addressed as an unconstrained loss minimization problem. To this end, we rewrite the BU-HTMM likelihood in eq.~(\ref{eq:SPBHTMMLik}) to highlight distribution parameters as follows,
\begin{equation}\label{eq:htmLogLik}
\begin{split}
    \log & \mathcal{L}_c(\mathbf{x}|\theta^{m}) =  \sum_{u' \in \mathcal{LF}} \sum_{i=1}^{C} z_{u'i} (\log \pi_i + \log b_{i}(x_{u'}))\\
    & + \sum_{u \in  \mathcal{U} \setminus \mathcal{LF}} \sum_{i,j = 1}^{C} \sum_{l = 1}^{L} z_{uijl} (\log \varphi_l + \log A_{ij}^{l} + \log b_{i}(x_{u})),
\end{split}
\end{equation}
where, for the sake of simplicity, we are providing the log-likelihood for a single sample $\mathbf{x}$, while subscript $m$ denotes the $m$-th generative model in the HTN (indicated only for $\theta$ for brevity). The formulation in eq.~(\ref{eq:htmLogLik}) is written in terms of:
\begin{itemize}
  \item $A_{ij}^{l}$, the probability of transiting to state $i$ given the $l$-th child in state $j$;
  \item $\pi_i$, the prior probability of a leaf being in state $i$;
  \item $b_{i}(v)$, the probability of state $i$ emitting node label $v$, and
  \item $\varphi_l$, the switching parent probability for the $l$-th child.
\end{itemize}
The terms $z_{ui}$ and $z_{uijl}$ are latent indicator variables for the node $u$ being in state $i$ and for the joint observation of node $u$ being in state $i$ while its $l$-th child is in state $j$, respectively (see \cite{DBLP:journals/tnn/BacciuMS12} for further details).

In order to achieve constraint-free optimization of (\ref{eq:htmLogLik}), we reparameterize the items in the list above using a softmax basis, i.e.
\[
A_{ij}^l =\frac{\exp \lambda_{ijl}^{(A)}}{\sum_i {\exp \lambda_{ijl}^{(A)}}}, \ \ \pi_{i} =\frac{\exp \lambda_{i}^{(\pi)}}{\sum_i {\exp \lambda_{i}^{(\pi)}}}
\]
\[
b_{i}(v) =\frac{\exp \lambda_{iv}^{(b)}}{\sum_v {\exp \lambda_{iv}^{(b)}}}, \ \ \varphi_{l} =\frac{\exp \lambda_{l}^{(\varphi)}}{\sum_l {\exp \lambda_{l}^{(\varphi)}}}
\]
where $\theta^{(m)} = \{\mathbf{\lambda}^{(A,m)}, \mathbf{\lambda}^{(\pi,m)}, \mathbf{\lambda}^{(b,m)}, \mathbf{\lambda}^{(\varphi,m)}\}$ are the new free parameters of the $m$-th generative model. Given this new formulation for the BU-HTMM, we can proceed with the derivation of the gradients needed to minimize (\ref{eq:loss}) with respect to $\theta=\{\mathbf{W}^o,\theta^{(1)},\dots,\theta^{(M)}\}$ by appropriate application of the chain rule for calculus. Sparing the details of tedious derivations, we obtain the following gradients for a labelled sample $(\mathbf{x},\mathbf{d})$:
\begin{equation}\label{eq:wgrad}
\frac{\partial L(\mathbf{x}|\theta)}{\partial W_{(m,r)k}^o} = -(d_k - p_k) h(\mathbf{x})_{(m,r)}
\end{equation}
\begin{equation}\label{eq:bugrad}
\frac{\partial L(\mathbf{x}|\theta)}{\partial \theta^{(m)}} = \sum_{k}\sum_{r} \frac{\partial L(\mathbf{x}|\theta)}{\partial W_{(m,r)k}^o} \frac{\partial h(\mathbf{x})_{(m,r)}}{\partial \theta^{(m)}} \frac{\partial \log \mathcal{L}_c(\mathbf{x}|\theta^{m})}{\partial \theta^{(m)}}
\end{equation}
where
\begin{equation}\label{eq:hgrad}
\frac{\partial h(\mathbf{x})_{(m,r)}}{\partial \theta^{(m')}} = \left( 1 - h(\mathbf{x})_{(m,r)}^2\right) \cdot (\delta_{m'm} - \delta_{m'r})
\end{equation}
using an indicator variable $\delta_{m'm} = 1$ when $m' = m$ and it is zero otherwise.

The derivative of the $m$-th model log-likelihood in eq.~(\ref{eq:bugrad}) is equal to its expected value $E[\log \mathcal{L}_c(\mathbf{x}|\theta^{m})]$ taken under the posterior distribution of its indicator variables $z_{ui},z_{uijl}$ (as in classical EM learning). This yields to the following parameter-specific gradients, completing model derivation:
\begin{equation}\label{eq:Ag}
\frac{\partial \log \mathcal{L}_c(\mathbf{x}|\theta)}{\partial \lambda_{ijl}^{(A)}} = \sum_{u \in  \mathcal{U} \setminus \mathcal{LF}} \left(\epsilon_{u,ch_l(u)}(i,j) - \epsilon_{ch_l(u)}(j) A_{ij}^l\right),
\end{equation}
\begin{equation}\label{eq:pig}
\frac{\partial \log \mathcal{L}_c(\mathbf{x}|\theta)}{\partial \lambda_{i}^{(\pi)}} = \sum_{u \in \mathcal{LF}} \epsilon_{u}(i)  - \pi_{i} \cdot |\mathcal{LF}|,
\end{equation}
\begin{equation}\label{eq:bg}
\frac{\partial \log \mathcal{L}_c(\mathbf{x}|\theta)}{\partial \lambda_{iv}^{(b)}} = \sum_{u \in \mathcal{U}} \left(\epsilon_{u}(i)(\tau_v(x_u) - b_{i}(v))\right),
\end{equation}
\begin{equation}\label{eq:phig}
\frac{\partial \log \mathcal{L}_c(\mathbf{x}|\theta)}{\partial \lambda_{l}^{(\varphi)}} = \sum_{u \in \mathcal{U} \setminus \mathcal{LF}} \left(\sum_{i,j}^C \epsilon_{u,ch_l(u)}(i,j) - |\mathcal{U} \setminus \mathcal{LF}| \varphi_l  \right),
\end{equation}
where $|\cdot|$ is set cardinality and $\tau_v(x)$ in eq.~(\ref{eq:bg}) is an indicator function equal to one when $v = x$ (and is zero otherwise). The latent state posteriors
\begin{equation}\label{eq:Ag}
\epsilon_{u,ch_l(u)}(i,j) = P(Q_u = i, Q_{ch_l(u)} = j, S_u = l | \mathbf{x})
\end{equation}
\begin{equation}\label{eq:Ag}
\epsilon_{u}(i) = P(Q_u = i | \mathbf{x})
\end{equation}
are computed using the standard E-step pass for the BU-HTMM, referred to as upwards-downwards algorithm: see \cite{DBLP:journals/tnn/BacciuMS12} for details. This last step concludes the derivation of the gradients for the HTN model, whose parameters can then be updated using any gradient-based algorithm, such as stochastic gradient descent (SGD), minibatch SGD, etc. In the following section, we show the experimental assessment of an HTN model based on a SGD update, i.e. for a generic parameter $\theta$ at time $t+1$ and a sample tree $\mathbf{x}$ (with associated classification $\mathbf{d}$) we have
\[
\theta^{t+1} = \theta^{t} + \nu^t \frac{\partial L(\mathbf{x}|\theta^{t})}{\partial \theta}
\]
where $\nu^t$ is an exponentially decaying learning rate.

\section{Experimental Results} \label{sect:expcomp}

\subsection{Data and Experimental Setup} \label{sect:setup}
We provide an experimental assessment of the HTN model on publicly available benchmarks on tree-data classification, spanning different application areas and including structures with different properties (e.g. tree outdegree, number of classes, sample size, size of label vocabulary, etc.). Table \ref{table:datasets} summarizes the main characteristics of the datasets used for this experimental assessment. The first benchmark concerns the classification of XML formatted documents from a corpus used in the 2005 INEX Competition \cite{inex2005}. This dataset is characterized by a large sample size and by a large number of unbalanced classes; trees are generally shallow, with a large outdegree. Standard splits into training and test sets are available for this dataset \cite{inex2005}, where roughly half of the total samples are used for training. The second set of benchmarks concerns the classification of the molecular structure of glycans, that can be represented by rooted trees where nodes stand for mono-saccharides and edges stand for sugar bonds. We consider two datasets from the KEGG/Glycan database \cite{kegg2004}, referred to as the Leukemia and Cystic data \cite{glyc49}. These benchmarks differs considerably from INEX: the task is binary and a small number of samples is available; trees are small and have a small outdegree. The third set of experiments deals with parse trees representing English propositions from a set of Dow-Jones news articles and associated semantic information. We employ a version of the Propbank dataset \cite{KingsburyPalmer02} introduced by \cite{AiolliPropbank}, that includes a sample from section $24$ of Propbank comprising $7,000$ training trees and $2,000$ validation examples, as well as  $6,000$ test samples extracted from section $23$ \cite{AiolliPropbank}. This benchmark defines a binary classification problem with a very unbalanced class distribution, where the percentage of positive examples in each set is roughly $7\%$.
\begin{table}[tb]
\caption{Summary of the datasets used in the experimental analysis \label{table:datasets}}
\begin{center} \footnotesize
\begin{tabular}{l|cccc}
  \hline
  Name & \# Trees & Classes & Outdegree & \# Labels\\
  \hline
  INEX 2005 \cite{inex2005} & $9361$ & $11$ & $32$ & $366$\\
  Leukemia \cite{glyc49} & $442$ & $2$ & $3$ & $57$\\
  Cystic \cite{glyc49} & $160$ & $2$ & $3$ & $29$\\
  Propbank \cite{AiolliPropbank} & $15000$ & $2$ & $15$ & $6654$\\
  \hline
\end{tabular}
\end{center}
\vspace{-5mm}
\end{table}

We have explored different HTN configurations by varying both the number of hidden tree Markov models $M$ as well as the number of hidden states $C$. Given the considerable differences in the sizing of the datasets and in the structural properties of their trees, we have considered different choices for the $M$ and $C$ hyperparameters depending on the benchmark at hand. In particular, for INEX 2005 and Propbank we have considered $C \in \{2, 4, 6, 8\}$ and $M \in \{20, 40, 60, 80\}$, while for the Glycans task we have tested $C \in \{2, 4, 8\}$ and $M \in \{10, 20, 30\}$. The number of tested hidden states has been determined following the guidelines in the original paper for the bottom-up Markov tree model \cite{DBLP:journals/tnn/BacciuMS12}.

The model hyperparameters $M$ and $C$ have been chosen in model selection: for INEX 2005, this has been performed through a 3-fold cross validation on the training set split, while for Propbank data we have used the predefined validation set. For the Glycans tasks, on the other hand, results have been obtained by a stratified 10-fold CV using the available standard partitions. For each fold, we have used different random initializations for the hidden Markov models distributions. Training of the HTN model has been performed through stochastic gradient descent as described in Section \ref{sect:approach},  with exponentially decaying learning rate (initial value $\nu = 0.01$ and maximum number of epochs equal to $100$) and Nesterov momentum (with initial weight $\alpha_0=0.5$ and final weight $\alpha_T=0.9$). The model-selected configuration for each dataset has then been evaluated on an external test set, except for the Glycans tasks where results in literature are provided as mean ROC-AUC on the 10-folds \cite{glyc49}.

\subsection{Results and Analysis} \label{sect:results}
Table \ref{tab:results} summarizes the results of the application of HTM to the datasets described in the previous section. Here, the predictive performance of the HTM model is compared with that of state-of-the-art generative tree kernels \cite{genKernel,icann12} built on the top of the same hidden Markov tree models used by HTM and denoted as JK in Table \ref{tab:results}. The classifiers for JK have been realized through support vector classification, using the publicly available LIBSVM \cite{libsvm01} software as described in \cite{genKernel}. A cross-validation procedure using the same validation scheme used by HTM has been applied to the generative kernels to select the number of hidden states $C$ and the value of the SVM cost parameter $C_{svm}$ from the following set of values: $0.001$, $0.01$, $0.1$, $1$, $10$, $100$, $1000$.
Additionally, Table \ref{tab:results} provides reference results from popular syntactic tree kernels in literature such as ST \cite{Vishwanathan2003}, SST \cite{sst}, PT \cite{Moschitti2006}.
\begin{table*}[tb]
  \centering
  \caption{Classification performance on the external (out-of-sample) test set comparing HTN with tree kernels on the configuration selected in cross-validation. Performance for the INEX task is expressed as classification accuracy (\%), for Propbank as F1 score and for the Glycans datasets as AUC-ROC. The best result for each dataset is highlighted in bold.} \label{tab:results}
  \begin{tabular}{|c||c|c||c|c||c||c||c||}
    \hline
    & \multicolumn{2}{|c||}{HTN} & \multicolumn{2}{c||}{JK} & ST & SST & PT\\
    Dataset & Configuration & Test & Configuration & Test & Test & Test & Test\\
    \hline
    INEX 2005 (Acc. $\%$) & $C = 8, M=60$ & $96.03$ & $C=8$ & $94.22$ & $88.73$ & $88.79$ & $\mathbf{97.04}$\\
    \hline
    Propbank (F1) & $C = 4, M=40$ & $\mathbf{0.699}$ & $C=10$ & $0.567$ & $0.51$ & $0.542$ & $0.516$ \\
    \hline
    Cystic (AUC)& $C = 2, M=20$ & $\mathbf{0.864}$ & $C = 10$ & $0.796$ & $0.798$ & $0.696$ & $0.823$\\
    \hline
    Leukemia (AUC)& $C = 2, M=20$ & $\mathbf{0.974}$ & $C = 8$ & $0.966$ & $0.961$ & $0.933$ & $0.967$\\
    \hline
  \end{tabular}
  \vspace{-2mm}
\end{table*}

Results highlight that HTM outperformed the reference kernels in most of the task, with the only exception of INEX 2005 where the PT kernel has still state of the art performance, while HTM is the runner-up model with a classification accuracy which is not far from that of PT. On the other hand HTM performance is significantly higher on the Propbank data, where it increases the F1 score by a notable $0.13$ points from the second best model, using many (i.e. $40)$ generative models of small complexity (i.e. only $C=4$ hidden states) as compared to the few generative models of high complexity ($C=10$) used by the JK.  Such a tendency to favour many smaller models over a few more complex HTMM is also evident on the Glycans task, where again HTM is the best performing model.

All in all, HTM seems able to challenge state-of-the-art syntactic kernels while exploiting the ability of generative tree models in capturing structural representations from data in a more effective way than the JK approach.  To better assess the effectiveness of HTM with respect to JK in exploiting the underlying generative models, we have performed an additional experiment involving INEX 2005. Here, we have been focusing on the tradeoff between classification accuracy and model complexity, defined in terms of number of number of hidden states as well as in terms of number of generative models employed. The choice of INEX 2005 is motivated by the non-trivial number of classes, i.e. $11$, which for the generative setting underlying JK means that this kernel is using $11$ HTMM to define its feature space. As for the HTM, instead, we have varied the number of generative models with $M \in \{6,11\}$ (here $6$ has been chosen as it is roughly half of the number of classes in the benchmark).
\begin{figure}[tb]
\includegraphics[width=1\columnwidth]{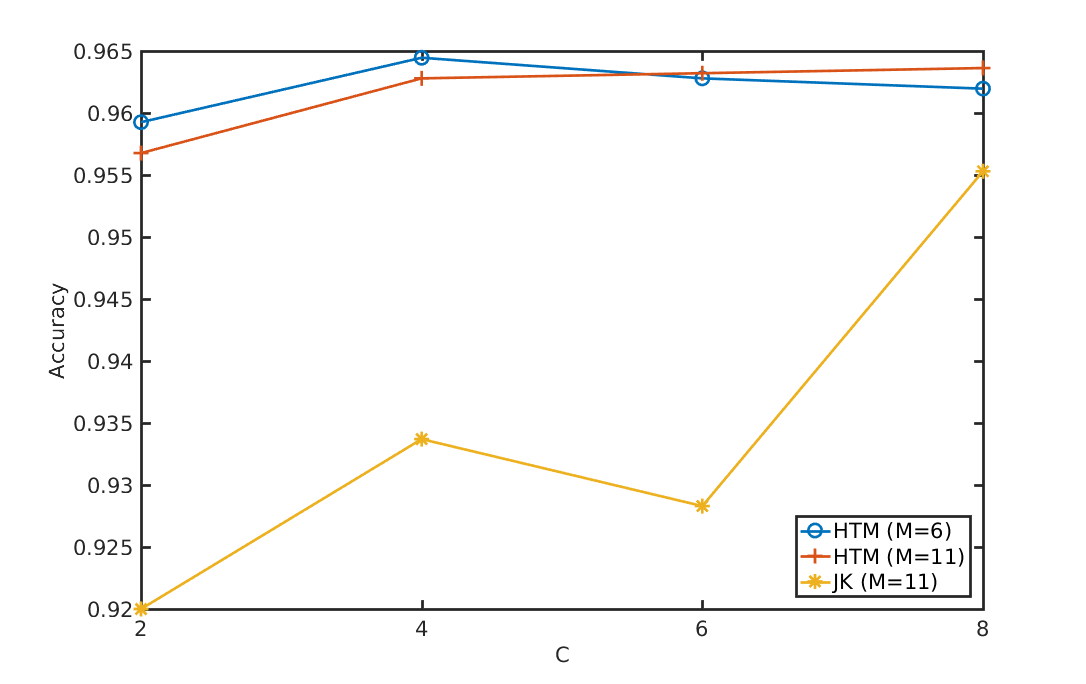}
\caption{Plot of the accuracy-complexity tradeoff on the INEX 2005 dataset for HTN compared with a JK using the same underlying generative model (test accuracy on the $y$-axis). The number of generative models is fixed to class number in JK (i.e. $11$), while the two curves for HTN denote results for $M \in \{6,11\}$. \label{fig:complex}}
\end{figure}

Figure \ref{fig:complex} shows the result of this analysis, showing a clear advantage of HTN in terms of classification performance over JK for all the configurations under test. In order to reach competitive performances, JK needs a larger hidden state space than HTN while using $M=11$ models. On the other hand, HTN is able to cope both with lesser models, c.f. its performance when only $M=6$ models are used, as well as when the generative models have reduced state spaces. In particular, HTN achieves competitive performances with as little as $C=2$ hidden states per generative model. This seems to confirm our initial intuition that by training the generative models as part of a deep neural architecture one can obtain more discriminatively-tuned structural feature detectors than by training the generative models in isolation and then fusing their contribution only at the classification stage, as in the JK approach.

\section{Conclusions} \label{sect:conclude}

We have introduced a deep learning approach for the adaptive processing of tree data based on a modular architecture comprising several parallel generative models that act as detectors of structural features. Such detectors are learned from data by incorporating discriminative information into the training of the hidden Markov tree models by backpropagation of the error generated by the neural output layer. A contrastive layer \cite{brown2002relative} of sparsely connected neurons with fixed binary weights serves to force differentiation of the feature detectors by confronting their predicted likelihoods and pushing them apart due to the inverted signs in the contrastive weights.

The experimental analysis highlights the ability of the HTN model in learning an encoding of the structural information that is effective for the target predictive task. In particular, experimental results show how HTN outperforms generative kernels based on the very same underlying generative models. Further, HTN can outperform state-of-the-art syntactic tree kernels.

HTN lays the basic architecture for integrating generative models with different representation capabilities into the same model, allowing to learn richer and more discriminative structural feature encodings. In particular, as shown by \cite{ijcnn2014}, the integration of information from top-down and bottom-up HTMM can yield to superior predictive performance over the two approaches in isolation. One ongoing extension of the model considers an HTN comprising both bottom-up and top-down models. Another ongoing development concerns the application of HTN to the processing of more complex and general classes of graphs. In fact, it is possible to use the modular structure of HTN to perform different parallel visits of the graph, using the neural layers to integrate the information extracted by the visits in the generative models. Finally, we are currently studying how the modular architecture of HTN can be exploited to parallelize parameters fitting on GPU cards using minibatching (source code for the GPU accelerated HTN will be released to the community\footnote{Source code link will be inserted here after publication}).

\section*{Acknowledgment}
This work has been supported by the Italian Ministry of Education, University, and Research (MIUR) under project SIR 2014 LIST-IT (grant n. RBSI14STDE). The Author gladly acknowledges Dell-EMC and Nvidia for donating the Dell C4130 server and the Nvidia M40 GPUs used to perform the experimental analysis.


\end{document}